\documentclass[letterpaper, 10 pt, conference]{ieeeconf}  

\IEEEoverridecommandlockouts                              

\usepackage[utf8]{inputenc}
\usepackage{authblk}
\usepackage{hyperref}
\usepackage{graphicx}
\usepackage{epsfig} 
\usepackage{amsmath} 
\usepackage{amsfonts}
\usepackage{subfig}
\usepackage{color}
\usepackage{glossaries}
\usepackage[]{algorithm2e}
\usepackage{soul}

\graphicspath{ {./} }

\title{WoLF: the Whole-body Locomotion Framework for Quadruped Robots}

\author[1]{Gennaro Raiola}
\author[1, 2]{Michele Focchi}
\author[3]{Enrico Mingo Hoffman}
\affil[1]{Istituto Italiano di Tecnologia, Via Morego, 30 16163 Genova, Italy}
\affil[2]{Universit\`a di Trento, Via Sommarive, 9, 38123 Trento, Italy}
\affil[3]{PAL Robotics, Carrer de Pujades 77, 08005 Barcelona, Spain}
\date{January 2022}

\begin{document}

\maketitle

\begin{abstract}

The Whole-Body Locomotion Framework (WoLF) is an end-to-end software suite devoted to the loco-manipulation of quadruped robots. 
WoLF abstracts the complexity of planning and control of quadrupedal robot hardware into a simple to use and robust software that can be connected through multiple tele-operation devices to different quadruped robot models.
Furthermore, WoLF allows controlling mounted devices, such as arms or pan-tilt cameras, jointly with the quadrupedal platform.
In this short paper, we introduce the main features of WoLF and its overall software architecture.
Paper Type -- Recent Work \cite{raiola2020}.
\end{abstract}
%

\section{Introduction}

Nowadays, a lot of emphasis and efforts have been devoted to autonomous 
exploration and patrolling of dangerous and hazardous environments using robotic systems. 
In particular, a promising type of platform is represented by quadruped robots.
Quadrupeds are becoming more and more interesting  mostly due to: 
\begin{itemize}
    \item the augmented mobility w.r.t. the wheeled/track counterparts,
    \item the higher stability and reliability than bipeds, due to a bigger support base,
    \item the increased payload w.r.t. drones, with the possibility to carry manipulator(s) that endow them with manipulation capabilities. 
\end{itemize}
%
%
These advantages make them particularly suited to explore and navigate cluttered and unstructured environments, at the cost of an increased complexity associated with the control layer, in particular concerning locomotion, that is often tailored for a specific robot. 
Complexity increases when the robot is equipped with one (or more) manipulator(s) to perform manipulation tasks.
In this case,  major difficulties arise due to the connection of two different robotics systems, and their synchronous control.
\par
While several companies are now launching on the market their own quadruped robots, e.g. Boston Dynamics~\footnote{\url{https://www.bostondynamics.com/}}, Unitree Robotics~\footnote{\url{https://www.unitree.com/}}, Anybotics~\footnote{\url{https://www.anybotics.com/}} and PAL Robotics~\footnote{\url{https://pal-robotics.com/}},  \textit{generic} and \textit{robot agnostic} control software solutions are not yet available. 
In the best of the cases, black box solutions are provided by some companies to realize high-level navigation for the platform, that cannot be accessed/modified by the user, thus strongly limiting their flexibility.   
\par
The absence of a common and standardized software framework, together with the high level of expertise required to control such platforms, prevents end-users to exploit the full potential of the market of quadruped robots.
\vspace{-0.3cm}
\begin{figure}[thb!]
    \centering
    \includegraphics[width=\columnwidth, trim={7cm 5.5cm 7cm 5.5cm}, clip=true]{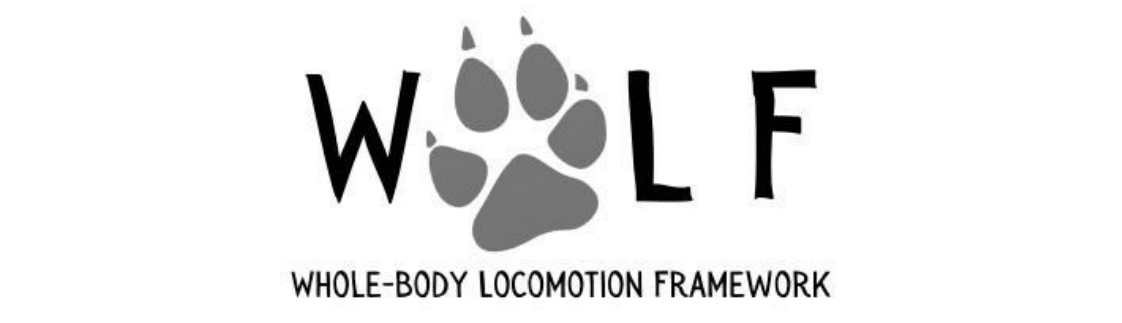}
    \caption{WoLF logo.}
    \label{fig:wolf_logo}
\end{figure}
\vspace{-0.3cm}
With WoLF (Figure~\ref{fig:wolf_logo}), we provide a plug-and-play software framework, easy to tune and adaptable to any quadruped robot without the need for specific knowledge about locomotion or control.
WoLF is based on established robotics tools and technologies such as ROS~\cite{quigley2009ros}, Gazebo~\cite{agueroVRC2015}, OpenSoT~\cite{hoffmanOpenSoT2017} and more, to promote standardization and ease of use. 
To be able to connect WoLF to different kinds of quadruped robots, we designed it to work as a plugin for ros\_control \cite{chitta2017ros_control}. 
The ros\_control package permits to easily abstract the particular hardware for both the quadruped and/or the manipulator and therefore to re-use the same controller with different robots provided that they expose an effort interface.
%
%
%
%
\par
We believe that WoLF could be useful for practitioners, companies, or research institutions that want to build and deploy their own solutions based on off-the-shelf quadruped platforms.
According to a report by Mondor Intelligence\footnote{\url{https://www.mordorintelligence.com/industry-reports/search-and-rescue-robots-market}}, the Search and Rescue Robots Market is projected to grow with a Compound Annual Growth Rate (CAGR) of more than 20\%.
Another source, Allied Market Research, reported that the global inspection,  and surveillance robots market generated \$940 million in 2020 and is expected to reach close to \$14 billion by 2030\footnote{\url{https://www.cnbc.com/2021/12/26/robotic-dogs-taking-on-jobs-in-security-inspection-and-public-safety-.html}}.
Other applications are disaster recovery, search \& rescue, decontamination (e.g. hazardous materials handling), maintenance, human-robot collaboration, and  space exploration.
\section{WoLF components}
\label{sec:components}

\begin{figure*}[h]
    \centering
    \includegraphics[width=\textwidth]{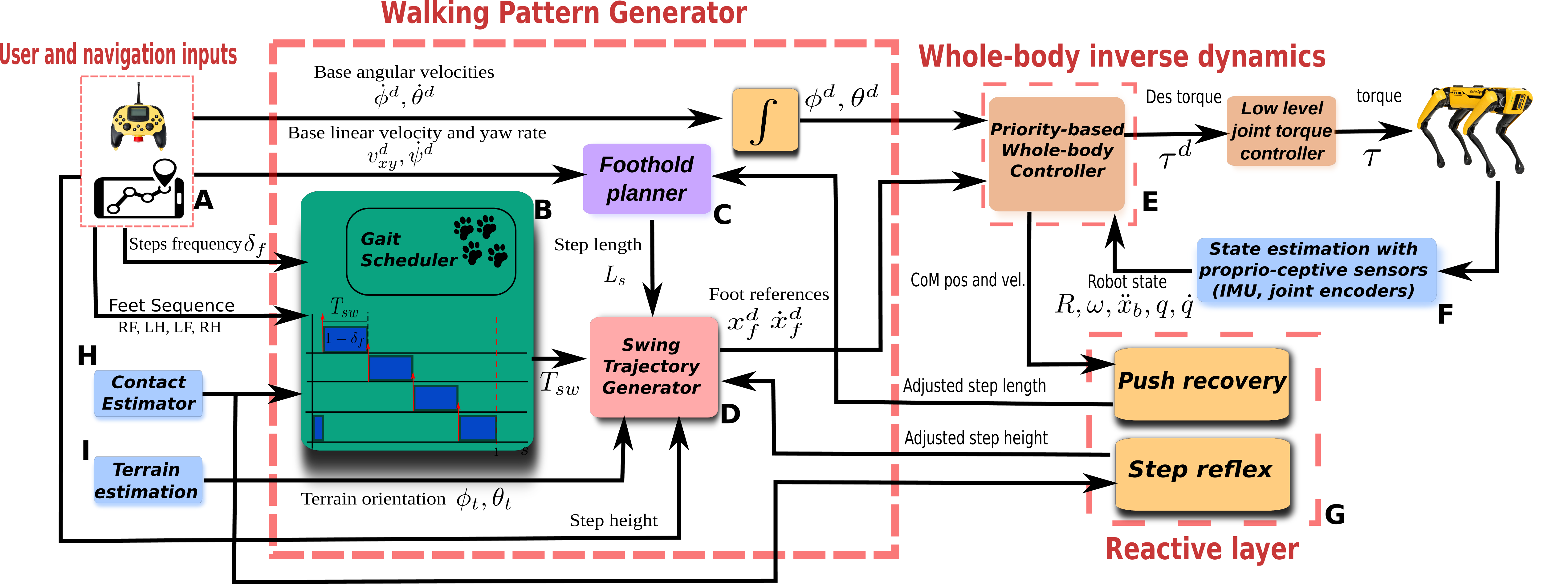}
    \caption{WoLF block diagram overview. The diagram is composed of four logical layers: the user and navigation input layer, the walking pattern generator, the whole-body inverse dynamics, and the reactive layer. Each layer is composed of one or more components. A description of the layers and their components is detailed in Section~\ref{sec:components}. }
    \label{fig:diagram}
\end{figure*}

The WoLF project started from the work in~\cite{raiola2020} about a novel locomotion framework for quadrupedal robots. 
In this seminal work a gait scheduler, a foothold planner and a whole-body controller was implemented. Starting from that,  WoLF has grown in terms of capabilities adding the possibility to control mounted devices in a whole-body manner, navigate and map the environment, support diverse input interfaces and, last but not least, incorporating reactive  strategies to increase robustness to disturbances/uncertainties. 

WoLF follows the same philosophy of CHAMP \footnote{\url{https://github.com/chvmp/champ}}, and  has been designed with the aim to support most of the off-the-shelf quadruped robots available on the market, making them easily integrable into the framework. Differently from CHAMP, WoLF features support for on-board manipulators. 
Furthermore, the framework has been designed with the main purpose to extend the capabilities of quadrupedal platforms with manipulation and navigation skills.
%
%

A general overview of the various components in WolF is shown in (Figure~\ref{fig:diagram}).

\subsection{User inputs and navigation}

The \emph{user interface} (Figure~\ref{fig:diagram} - A) is the layer exposed to the user and to the external ROS world, where multiple control devices (joy-pads, keyboards, GUI, ROS topics and services, etc...) can be easily connected to have full control of the robot. 
%
%
%
WoLF provides a ROS topic interface to send desired twist commands to the robot's base which can be connected to the ROS package move\_base\footnote{\url{http://wiki.ros.org/move_base}} that represents the entry point for the ROS navigation stack. 
The move\_base package links together a global and local planner to accomplish global navigation tasks such as moving the robot to a specific point on the map.

In order to allow shared-autonomous navigation, the twist interface works as a low priority command interface, which means that at any moment the operator can take full control of the robot by using the selected input device (joy-pad, keyboard, spacemouse, etc...).
An essential part of the navigation stack is the Simultaneous Localization And Mapping algorithm (SLAM). 
With WoLF, we opted to use hector SLAM \cite{hector2011}: one of the most used and reliable algorithms for SLAM in robotics. 
Thanks to the modularity of the ROS navigation stack and the underlying ROS communication system, it is possible to easily integrate WoLF with different SLAM algorithms if needed.

\subsection{Walking pattern generator and reactive layer}
The walking pattern generator is one of the main layers of WoLF. Its purpose is to transform input commands coming from the user or from the navigation stack into references for the whole-body inverse dynamics layer. It is composed of the following components:

\begin{itemize}
\item gait scheduler (Figure~\ref{fig:diagram} - B): it  coordinates the footsteps based on the gait schedule,
\item foothold planner (Figure~\ref{fig:diagram} - C): it transforms base twist commands into footholds,
\item swing trajectory generator (Figure~\ref{fig:diagram} - D): it calculates the swing trajectory for the feet based on the desired footholds and the estimated terrain slope.
\end{itemize}
The reactive layer, instead, is composed of the push recovery and the step reflex components (Figure~\ref{fig:diagram} - G).
The push recovery is based on the mathematical definition of Instantaneous Capture Point (ICP) \cite{pratt2006}. 
When a push is detected, a delta for the footsteps is computed based on the ICP formulation. 
From a practical point of view, a push event is determined when the CoM of the robot gets closer to its support polygon boundary. 
The sensitivity to the push event  can be adjusted  by scaling the support polygon by a scalar value between 0 and 1. 
The step reflex, instead, is based on the work presented by Focchi et al. \cite{focchi2013local}. When a foot is swinging in the air, if a contact is detected during the first half of the swing trajectory, a stepping reflex is triggered and it generates a new swing motion over the nominal trajectory to achieve a more stable foothold.

\subsection{Whole-Body Inverse Dynamics}
The whole-body inverse dynamics layer (Figure~\ref{fig:diagram} - E) is in charge to track user/navigation inputs and walking pattern generator references and transforming them into torque commands at the joint level.
This layer is based on hierarchical Whole-Body Inverse Dynamics solved using Quadratic Programming (QP) optimization.
In particular, our formulation can handle as well hardware limitations such as joint position, velocity, acceleration, and torque limits which are implemented as inequality constraints in the QP optimization.
\par
The same QP optimization infra-structure can be used to handle the presence of one or multiple arms attached to the robot's base in a modular plug-and-play fashion.
The user can specify in which frame to control the arms and if the control is, therefore ``whole-body'' or not (i.e. if the movement of the arms affects or not the movements of the base).
For the ease of use, we define two \emph{control modes}: \emph{WALKING} in which the arms are controlled w.r.t. the base of the robot and \emph{MANIPULATION}, in which the arms are controlled w.r.t. the base footprint. The first mode is useful to reduce the effects of the arm movements when the robot is moving (improving the balancing of the robot) while the second mode fully exploits the whole-body architecture to increase the workspace of the arms. In both cases, the balance recovery is used to guarantee stability while using the arms.
The user can interact with the arms using any input device such as a joystick, keyboard, interactive markers, and so on. 
In particular, the user sets reference poses for the controlled frame or motion way-points, that are interpolated using minimum jerk polynomials.
\subsection{State and Terrain Estimation}
The state estimation (Figure~\ref{fig:diagram} - F) uses proprioceptive sensors (joint positions and velocities) and the Inertial Measurement Unit (IMU) to compute the actual twist of the floating-base  of the robot. 
In particular, the formulation used in WoLF considers the so-called \emph{Horizontal Frame}, namely a frame that follows the base of the robot while its orientation is kept parallel to the ground.
\par
With this, the orientation of the \emph{Horizontal Frame} frame is given by the IMU, while floating base linear and angular velocities are computed using a QP from joint velocities, knowing the active contacts with the environment.
The status of the contacts can be computed from joint torque sensors, or with contact sensors if these are available in the platform (Figure~\ref{fig:diagram} - H).
\par
To guarantee stability while walking, the support polygon is updated at each stance (i.e. when all the feet are in contact with the ground) and its center is used to generate a reference for the CoM task. Therefore, the robot will tend to keep the CoM close to the center of the support polygon. Instead, the CoM velocities are calculated based on the input reference velocities so that the base of the robot follows the desired input velocity.
The terrain estimation (Figure~\ref{fig:diagram} - I) is performed by fitting a plane over the feet at stance. The estimation can be used to re-orient the friction cones \cite{caron2015stability} and the swing trajectories such that the robot can climb up ramps and stairs \cite{focchi2020heuristic} without scuffing.

\section{Software packages}

\begin{figure}[thb!]
    \centering
    \includegraphics[width=\columnwidth]{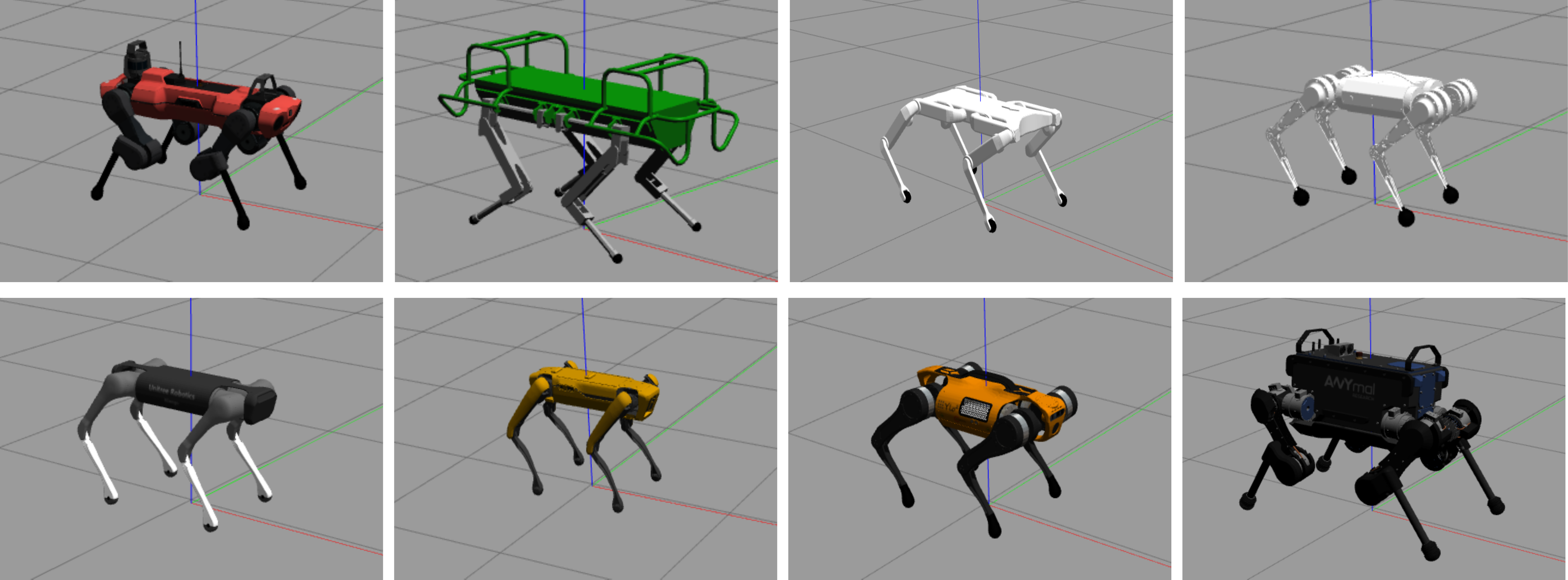}
    \caption{Robots currently supported in simulation by WoLF. From top-left to bottom-right: AnymalC, HyQ, Solo, Minicheetah, Aliengo, Spot\textsuperscript\textregistered \ , Ylo2, Anymal.}
    \label{fig:robots}
\end{figure}

The various WoLF packages are hosted on Github. The entry point to set up and run WoLF on any Ubuntu PC is its setup package\footnote{\url{https://github.com/graiola/wolf-setup}}. The other packages are:
\begin{itemize}
\item wolf\_descriptions\footnote{\url{https://github.com/graiola/wolf_descriptions}}: it contains robot and sensor descriptions used within the framework (Figure~\ref{fig:robots}). It is possible to use this package to add and try new robots.
\item wolf\_gazebo\_resources \footnote{\url{https://github.com/graiola/wolf_gazebo_resources}}: it contains Gazebo models and other resources to adapt and create customized simulation environments.
\item wolf\_hardware\_interface \footnote{\url{https://github.com/graiola/wolf_hardware_interface}}: it implements the hardware interface for ros\_control to be used with WoLF.
\item wolf\_gazebo\_interface \footnote{\url{https://github.com/graiola/wolf_gazebo_interface}}: This is the Gazebo hardware interface for ros\_control.
\item wolf\_aliengo\_interface \footnote{\url{https://github.com/graiola/wolf_aliengo_interface}}: Aliengo hardware interface for WoLF. 
\item wolf\_ylo2\_interface \footnote{\url{https://github.com/graiola/wolf_ylo2_interface}}: Ylo2 hardware interface for WoLF.
\item wolf\_navigation \footnote{\url{https://github.com/graiola/wolf_navigation}}: This package is used to provide navigation capabilities to WoLF. It integrates and provides several utilities such as odometry computation, way-point definition, and so on.
\end{itemize}

\section{Applications}

\begin{figure}[thb!]
    \centering
    \includegraphics[width=\columnwidth]{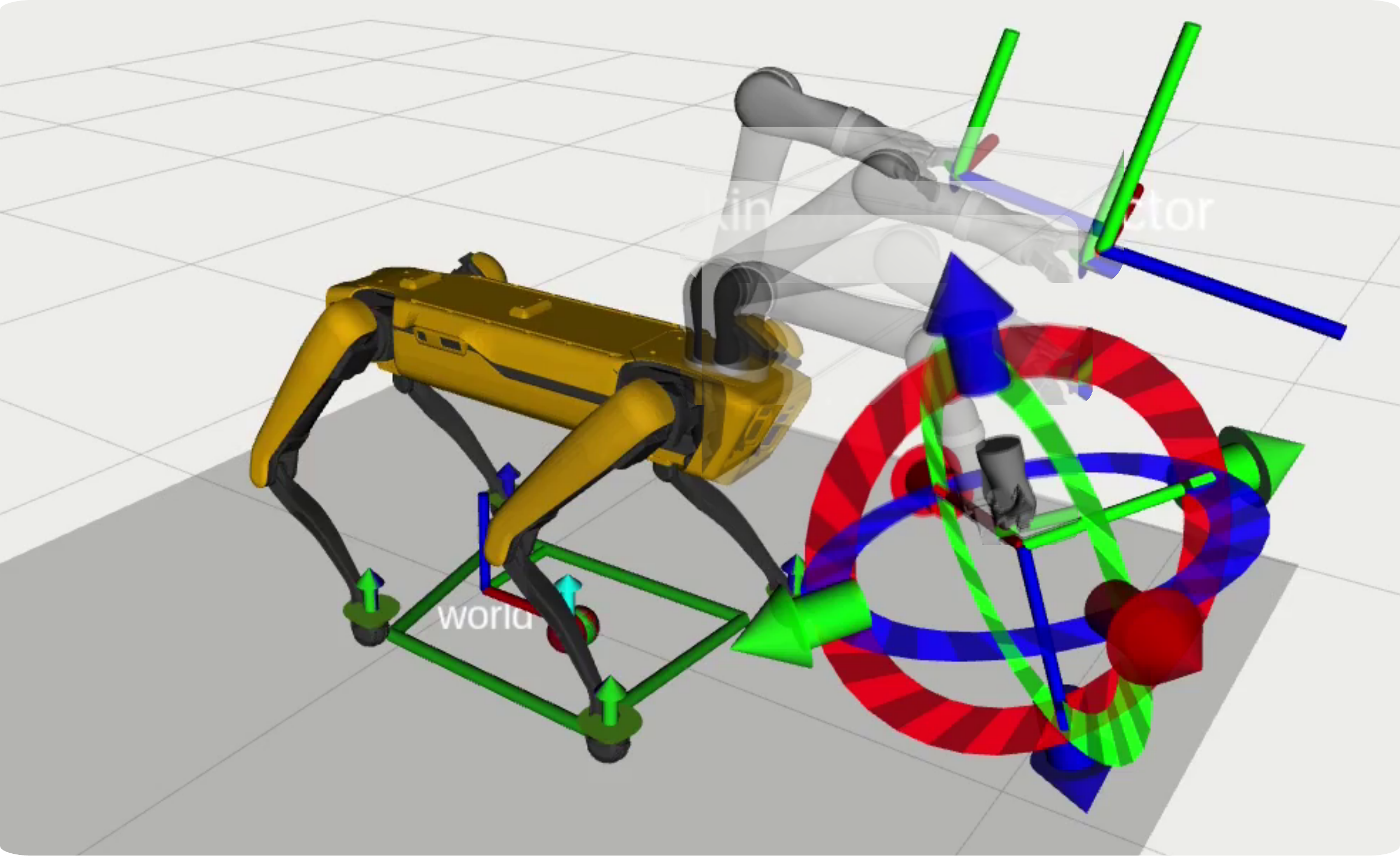}
    \caption{Spot\textsuperscript\textregistered \  with a kinova manipulator mounted on its base. WoLF permits to easily combine quadruped platforms with different robotic manipulators. In this example, the kinova end-effector is tele-operated with a ROS interactive marker.}
    \label{fig:spot_arm}
\end{figure}

In this section, we list some of the possible applications of the framework to use cases that are nowadays of increasing interest for the end-users. 
\begin{figure}[thb!]
    \centering
    \includegraphics[width=\columnwidth]{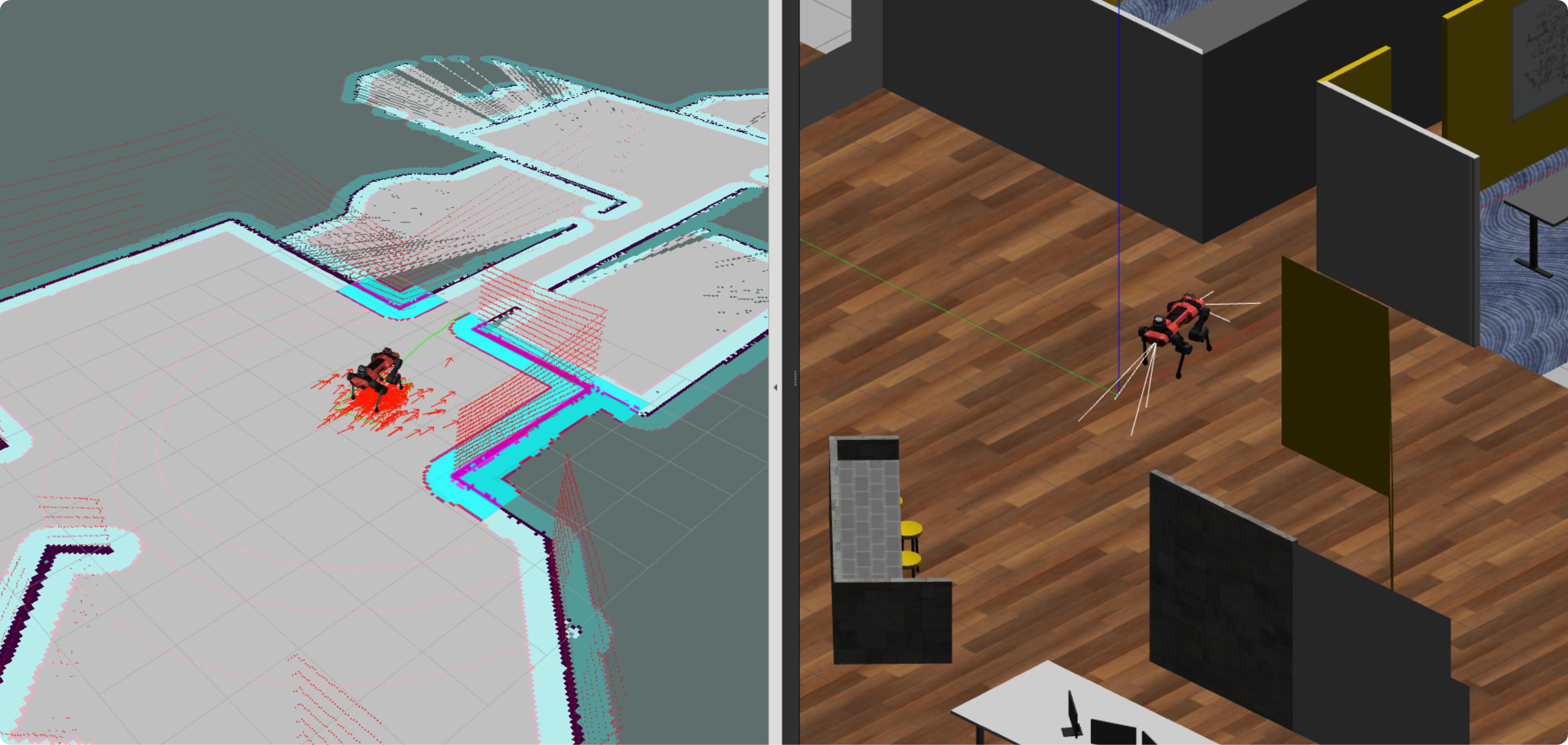}
    \caption{AnymalC navigating and reconstructing the map in simulation scenario.}
    \label{fig:anymalc_navigation}
\end{figure}

Applications range from nuclear decommissioning to mining, search \& rescue, inspection, and surveillance. 
In addition, this technology can be applied to flank human workers in order to reduce labor accidents, as well as in elderly care and space exploration.
The main focus for most end-users is the ability to  operate either autonomously or semi-autonomously, through tele-operation.
\par
An accompanying video\footnote{\url{https://youtu.be/O6TSUHiwSlU}} illustrates the main features of the WoLF framework showing omni-directional locomotion on uneven terrains, whole-body control with an additional mounted manipulator, navigation in human-structured environments, ramp and stair climbing with different quadruped platforms. 

\section{Conclusion and Future Works}
This short paper presented WoLF, the Whole-Body Locomotion Framework, an end-to-end software that simplifies loco-manipulation, mapping, and navigation in quadrupedal robotics systems.
WoLF is based on standard robotics tools such as ROS and ros\_control, making it easy to integrate on existing platforms and tele-operation devices.
This allows as well to seamlessly integrate different devices on the quadrupedal platform, such as manipulators or pan/tilt cameras.
\par
The interested reader can try out WoLF on its own machine via the GitHub repository\footnote{\url{https://github.com/graiola/wolf-setup}}. 
\par

The work has been preliminary tested on a Unitree Aliengo quadruped platform, to assess if the computational demand was compatible with a real hardware. We were able to run the framework at 1 kHz on an external Quad Core i7 laptop while sending torque commands to the robot.
Future works will regard the possibility to perform more dynamic motions such as jumps,  improvements on the loco-manipulation capabilities (when mounting a manipulator), as well as experiments on the real platforms.
%

\section*{ACKNOWLEDGMENT}
We would like to express our gratitude to Federico Rollo, Andrea Zunino, Ilyass Taouil, and Vincent Foucault for their valuable help in testing and integrating new features in WoLF.

\bibliographystyle{IEEEtran}	
\bibliography{biblio}

\end{document}